\documentclass{article}

\usepackage{PRIMEarxiv}

\usepackage[utf8]{inputenc} 
\usepackage[T1]{fontenc}    
\usepackage{hyperref}       
\usepackage{url}            
\usepackage{booktabs}       
\usepackage{amsfonts}       
\usepackage{nicefrac}       
\usepackage{microtype}      
\usepackage{lipsum}
\usepackage{fancyhdr}       
\usepackage{graphicx}       
\usepackage{cite}
\usepackage{amsmath,amssymb,}
\usepackage{algorithmic}
\usepackage{textcomp}
\usepackage{subcaption}
\usepackage{diagbox}
\usepackage{makecell}
\usepackage{array}
\usepackage{float}
\usepackage{enumitem}
\usepackage{multirow}
\graphicspath{{media/}}     
\newcolumntype{P}[1]{>{\centering\arraybackslash}m{#1}}
\title{Continuous Wavelet Transform and Siamese Network-Based Anomaly Detection in Multi-variate Semiconductor Process Time Series
}

\author{
  Bappaditya Dey$^*$, Daniel Sorensen$^*$, Minjin Hwang$^*$, Sandip Halder \\
  IMEC, Kapeldreef 75, 3001 Leuven, Belgium\\
  Correspondence: \texttt{Bappaditya.Dey@imec.be} \\ \\
  $^*$These authors contributed equally
}

\begin{document}
\maketitle

\begin{abstract}
Semiconductor manufacturing is an extremely complex and precision-driven process, characterized by thousands of interdependent parameters collected across diverse tools, process steps, and time scales. multi-variate time-series analysis (MTSA) has emerged as a critical methodology for enabling real-time monitoring, fault detection, and predictive maintenance in such environments. However, applying MTSA for anomaly prediction in semiconductor fabrication presents several critical challenges. These include the high dimensionality of sensor data, severe class imbalance due to the rarity of true faults, the presence of noisy and missing measurements, and the non-stationary behavior of production systems driven by dynamic recipe adjustments, tool aging, and maintenance activities. Furthermore, the complex interdependencies between process variables and the delayed emergence of faults across downstream stages significantly complicate both anomaly detection and root-cause-analysis. This paper presents a novel and generic approach for anomaly detection in multi-variate time-series data using machine learning, with a primary focus on semiconductor manufacturing processes. The proposed methodology consists of three main steps, as: a) converting multi-variate time-series (MTS) data into image-based representations using the Continuous Wavelet Transform (CWT), b) developing a multi-class image classifier by fine-tuning a pretrained VGG-16 architecture on custom CWT image datasets, and c) constructing a Siamese network composed of two identical sub-networks, each utilizing the fine-tuned VGG-16 as a backbone with shared weights. The network takes pairs of CWT images as input—one serving as a reference or anchor (representing a known-good or non-anomalous process/tool trace), and the other as a query (representing an unknown or potentially anomalous trace). The model then analyzes and compares the embeddings of both inputs to determine whether they belong to the same class at a given time step. Our proposed approach demonstrates high accuracy in identifying anomalies on a real FAB process time-series dataset, offering a promising solution for offline anomaly detection in process and tool trace data. Moreover, the approach is flexible and can be applied in both supervised and semi-supervised settings.
\end{abstract}


\section{Introduction}
In the semiconductor industry, node-technology has steadily advanced toward ever smaller node sizes, evolving from a few $\mu m$ just a few decades towards sub-30nm pitches for 5 nm node and below in current production. As pitch sizes shrink and device complexity grows, there is an ever-greater need for precise process control, accurate metrology and data analysis, and advanced defect inspection. These improvements are necessary to comply with the increasingly stringent tolerances in chip-fabrication. Ensuring tool stability is therefore essential to sustain both production quality and throughput. Even slight drifts or deviations from target process can produce defective devices or components. Moreover, any sudden hardware malfunction, beyond what predictive maintenance can anticipate, requires immediate intervention, leading to unplanned tool downtime and potential production halts. Both scenarios impose significant costs on semiconductor manufacturers and their production lines and, more broadly, on any manufacturing operation. As an indicative estimate, a flow irregularity in a cleanroom environment can result in losses ranging from \$500K to \$1M per batch of scrapped wafers \cite{FlowErrorCosts}. Similarly, unplanned equipment downtime can cost between \$100K and \$2M per hour \cite{DowntimeCosts}, depending on the industry and severity of the malfunction. In the first scenario, process drifts and anomalies should be predicted or detected in advance and corrected in real time. In the second scenario, tool-related issues, such as sensor failures or hardware malfunction, should also be anticipated ahead of time to allow for informed decision-making. Early detection can help flag such anomalies before wafer processing begins, potentially preventing wafer loss by halting operations in advance and thus reducing production costs. However, determining how early such predictions must be made--and identifying the optimal number of time-stamp points or the appropriate time window, whether in milliseconds or seconds--remains an active area of research.

The state of a (semiconductor FAB) tool at any given time can be characterized by its configurable parameters (such as valve positions, nozzle settings, electrical biases, and gas flow rates) together with sensor measurements from the process chambers (e.g. pressure, temperature, and gas-species concentrations). During operation, the tool continuously records these values at fixed time-intervals, resulting in what is known as multi-variate time series (MTS) data, that captures the dynamic behaviour of the tool. This data format is fundamental to support critical tasks such as anomaly detection and prediction of process performance metrics (for example, etch rate, deposition rate, or chemical-mechanical polishing rate). Machine learning (ML) researchers have been studying MTS analysis for many years \cite{Lapedes1987}. Prior to the widespread adoption of ML techniques, traditional statistical models, such as Autoregressive Integrated Moving Average (ARIMA) \cite{ARIMA} and Autoregressive Conditional Heteroscedasticity (ARCH) \cite{ARCH}, were commonly used. However, these models assume linear dependencies and often fail to capture the complex non-linear dynamics prevalent in manufacturing process data. To address these limitations, researchers have adopted more expressive ML models, beginning with simple Multi-Layer Perceptrons (MLPs) \cite{MLP} and evolving toward advanced architectures such as deep Convolutional Neural Networks (CNNs) \cite{TCN} and Recurrent Neural Networks, particularly Long Short-Term Memory (LSTM) \cite{LSTM} networks, which are capable of learning intricate temporal and spatial patterns from data. Additionally, various auto-encoder based frameworks \cite{SAEexample} have been employed to learn compact representations of normal tool and process behaviour, aiding in feature extraction and anomaly detection.

Although statistical models and machine learning algorithms have been applied to time series analysis for over four decades, their use on multi-variate time series (common in semiconductor tool data) remains under-explored due to several key challenges, as summarized in Table \ref{Tab:KeyChallenges}.

\begin{table}[h]
\small
\resizebox{\textwidth}{!}{
\begin{tabular}{|P{3cm}|P{4cm}|P{4cm}|P{4cm}|P{4cm}|}
\hline
\backslashbox{Challenge}{Feature} & 1 & 2 & 3 & 4 \\
\hline
\textbf{Data-centric} & 
\textbf{High-dimensionality} \newline 
\begin{itemize}
    \item Hundreds to thousands of time-synchronized process parameters from tools
    \item Curse of Dimensionality
    \item High risk of erroneous correlation
\end{itemize} & 
\textbf{Imbalanced dataset with sparse anomalies} \newline 
\begin{itemize}
    \item True faults/anomalies are stochastic and rare  \newline($<1\%$)
    \item Sparse faults (like drifts, degradation) occur subtly rather than abruptly
    \item Missing and noisy dataset (missing timestamps, incomplete wafer traces) may mask anomalies
\end{itemize} & 
\textbf{Data Heterogeneity} \newline 
\begin{itemize}
    \item Variability across tools/ chambers/recipes/wafers
    \item Individual independent/ dependent shifts can be introduced in feature distributions
\end{itemize} & 
\textbf{Non-stationary} \newline 
\begin{itemize}
    \item Evolving process characteristics with time due to recipe tuning or maintenance
    \item Lacks adaptive models to capture distribution shifts
\end{itemize} \\
\hline

\textbf{Model-centric} & 
\textbf{Temporal dependencies} \newline 
\begin{itemize}
    \item Needs memory-intensive models like LSTMs, attention-based models, etc.
    \item Anomalies may demonstrate as long-range dependencies across time
\end{itemize} & 
\textbf{Multi-variate interdependencies} \newline 
\begin{itemize}
    \item Inter-variable causal vs correlational links to be learned, which is difficult
    \item Faults in one sensor may cascade or be redundant due to tightly coupled variables
\end{itemize} & 
\textbf{Unknown anomaly} \newline 
\begin{itemize}
    \item Ground-truth for anomalies is often unavailable, unknown, or unreliable
    \item Requires unsupervised learning rather than supervised
    \item Requires explainable-AI to demonstrate what caused an anomaly, which variable(s) are critical, and at which particular time-step the anomaly happened
\end{itemize} & 
\textbf{Real-time detection} \newline 
\begin{itemize}
    \item Requires (near) real-time detection rather than offline and continual learning inference
\end{itemize} \\
\hline

\textbf{Domain-specific} & 
\textbf{Recipe changes and dynamic control} \newline 
\begin{itemize}
    \item Recipe changes and dynamic control (even within runs) may create concept drift
\end{itemize} & 
\textbf{Tool-to-tool and chamber matching} \newline 
\begin{itemize}
    \item Same process may behave differently across different tools/chambers
    \item Requires a generalized model to capture tool/chamber variability as domain shifts
\end{itemize} & 
\textbf{Multi-Stage Process Complexity} \newline 
\begin{itemize}
    \item Semiconductor manufacturing involves multi-process steps such as litho$\rightarrow$ etch$\rightarrow$ clean$\rightarrow$ deposition$\rightarrow$chemical mechanical polishing, etc.
\end{itemize} & \\
\hline

\textbf{Operational and Integration} & 
\textbf{Human-in-the-loop dependency} \newline 
\begin{itemize}
    \item Process experts and hardware engineers only validate anomaly triggers
\end{itemize} & 
\textbf{Intellectual-Property constraints} \newline 
\begin{itemize}
    \item Model generalizability can suffer due to restrictions on confidential data sharing from fabs/vendors
\end{itemize} & & \\
\hline
\end{tabular}}
\vspace{0.2cm} 
\caption{Summary of key challenges of modelling multi-variate time series in the semiconductor manufacturing domain.}
\label{Tab:KeyChallenges}
\end{table}

In this research, we present a novel deep learning based method for anomaly prediction in semiconductor processes by addressing the challenges of multi-variate time series analysis with numerous interconnected sensors, process steps, and tool parameters over time. The main contributions of our research are the following:
\begin{enumerate}
    \item We proposed a novel preprocessing method to convert raw multi-variate time series data into a time–frequency representation. Specifically, we applied the Continuous Wavelet Transform (CWT) \cite{Torrence1998} to transform each time-series signal into a 2D time–frequency image. Unlike the Fast Fourier Transform (FFT) \cite{Brigham1967}, CWT provides localized time-frequency information, making it particularly effective for detecting transient anomalies and subtle variations in process dynamics. The core motivations behind using CWT include:
    \begin{enumerate}
        \item identifying short-lived, localized anomalies that are difficult to capture in raw time-series data,
        \item capturing both high-frequency (fast fluctuations) and low-frequency (slow drifts) components, and
        \item transforming variable-length signals into fixed-size 2D representations suitable for CNN-based image classification.
    \end{enumerate}
    \item We developed a deep Convolutional Neural Network (CNN) model, primarily by fine-tuning a pre-trained VGG-16 architecture, to classify the CWT images into multiple process state classes. The CNN effectively learns spatially localized frequency patterns -- such as bursts, transients, oscillations, and shifts in frequency bands -- through its convolutional filters. These patterns represent meaningful edge-like time–frequency changes and tool-specific dynamic behaviors. The compact embeddings extracted by the CNN were then used to classify process/tool states into normal categories such as H\_L (High-to-Low), L\_H (Low-to-High), and OoB (Out-of-Box or sensor idle state), and abnormal ones such as L\_H\_R (Low-to-High with right-shift) and H\_L\_L (High-to-Low with left-shift) etc.
    \item We further developed a Siamese network consisting of two identical sub-networks, each leveraging the previously fine-tuned VGG-16 as its backbone with shared weights. The network takes as input pairs of CWT images, one as a reference or anchor (representing a known-good, non-anomalous process/tool trace) and the other as a query (representing an unknown or potentially anomalous trace). The comparison is conducted over sliding time windows from the start to the end of the time series. The network analyzes the feature embeddings from both branches and computes a similarity score to determine whether they belong to the same process state class. A low similarity score indicates a potential anomaly in the incoming trace relative to the known-good reference at that particular time step. We validated our approach on a real FAB process time-series dataset, achieving high accuracy (as 20-way cross-validation resulted in 100\% correct anomaly identification).
    
\end{enumerate}
This demonstrates the effectiveness of our proposed method for offline anomaly detection and localization in semiconductor process and tool trace data. Additionally, the proposed framework is scalable to a large number of variables/sensors and supports both supervised and semi-supervised configurations, making it a versatile solution for advanced process monitoring in semiconductor manufacturing.

\section{Related Work}
With the growing interest in smart monitoring of manufacturing tools in the semiconductor industry, several approaches and models have been taken to predict and detect anomalies. These approaches typically fall into one of three categories \cite{ZamanzadehDarban2024}: 1) \textbf{forecasting-based} approaches predict future values of a time series, such as process or tool parameters, using a preceding window of historical sensor data. Anomalies are identified by comparing the predicted values with actual measurements, which represent known normal behaviour; significant deviations from these expected values may indicate abnormal tool behaviour or process drift; 2) \textbf{reconstruction-based} approaches also employ sliding windows to learn a low-dimensional latent space representation of normal time-series segments. During training, the model is optimized to reconstruct the original signal from this representation. At inference time, the trained model attempts to reconstruct new signals, and if it fails to do so accurately, the discrepancy is treated as an anomaly. Large reconstruction errors, when compared to the baseline of known normal patterns, indicate potential abnormalities caused by equipment faults, recipe deviations, or other process-related issues; 3) Lastly, \textbf{representation-based} approaches aim to apply models (typically, self- or semi-supervised learning techniques) to latent space representation of the time series data. The objective is to develop a robust understanding of normal patterns across processes, recipes, or sensors signals by capturing the complex temporal and contextual correlations. Anomalies in new observations are identified as deviations from this learned representation, enabling the detection of subtle or previously unseen failure modes.

In the context of multi-variate time series (MTS) anomaly detection in the semiconductor industry, several methodologies have been proposed. Notably, Liao, D. et al. \cite{Chen2020} implemented a reconstruction-based approach using a stacked autoencoder framework, deploying two autoencoders per sensor (one operating in the time domain and the other in the frequency domain) within a chemical vapour deposition tool. The model detected anomalies by observing large mean squared errors between the reconstructed signals and the actual sensor readings. 

Mellah, S. et al. \cite{Mellah2022} implemented a representation-based approach by applying Independent Component Analysis (ICA) to extract the most informative features from MTS data. These features were then used as input to decision-tree based ensemble models for anomaly detection and classification. The model was evaluated on simulated sensor data designed to resemble real production variables, with 28.6\% of the data labeled as faulty. This approach achieved an F-measure of 99.8\% for anomaly classification.

Baek, M. and Kim, S. \cite{Baek2023} transformed sliding windows of time series into a signature matrix, which was input to a Convolutional Autoencoder (CAE) in order to detect anomalies in the data. For data classified as anomalous, a residual matrix was calculated and used as input to a MLP to predict replacement segments for the anomalous parts. Finally, the KernelSHAP algorithm was employed to identify the key contributing factors behind the replacement segments. This architecture achieved classification accuracies generally above 90\% and provided a degree of interpretability regarding the causes of the anomalies. 

In the research by Hwang, R. et al. \cite{Hwang2023}, a Long Short-Term Memory Autoencoder (LSTM-AE) was combined with a Deep Support Vector Data Description (SVDD) objective function. The proposed framework includes two autoencoders: first a LSTM-AE was used to pre-train the input data and extract compact representation; then a dense layer AE was trained using a loss function derived from the SVDD objective. This SVDD-based loss encourages the latent representations of normal data to lie within a hypersphere in latent space, while anomalies are mapped outside of it. Using this approach, outliers were successfully identified in 2 out of 15 processes. Although no significant anomaly patterns were found in the remaining processes, the two flagged processes revealed instability in their corresponding chambers, as indicated by the high number of detected anomalies. Further analysis indicated that the anomalies in these chambers were caused from a similar type of malfunction.

The remaining structure of this paper is organised as follows: Section \ref{Sec:Methodology} outlines the proposed methodology, including data preprocessing, anomaly induction and model training. Section \ref{Sec:R_D} presents the experimental results along with a detailed analysis. Section \ref{Sec:Limitations} outlines the key limitations of the current work and explores potential directions for future research. Finally, Section \ref{Sec:Conclusion} concludes the paper. 

\section{Methodology}
\label{Sec:Methodology}
\subsection{Data}
\begin{figure*}[htbp]
\centerline{\includegraphics[width=0.9\textwidth]{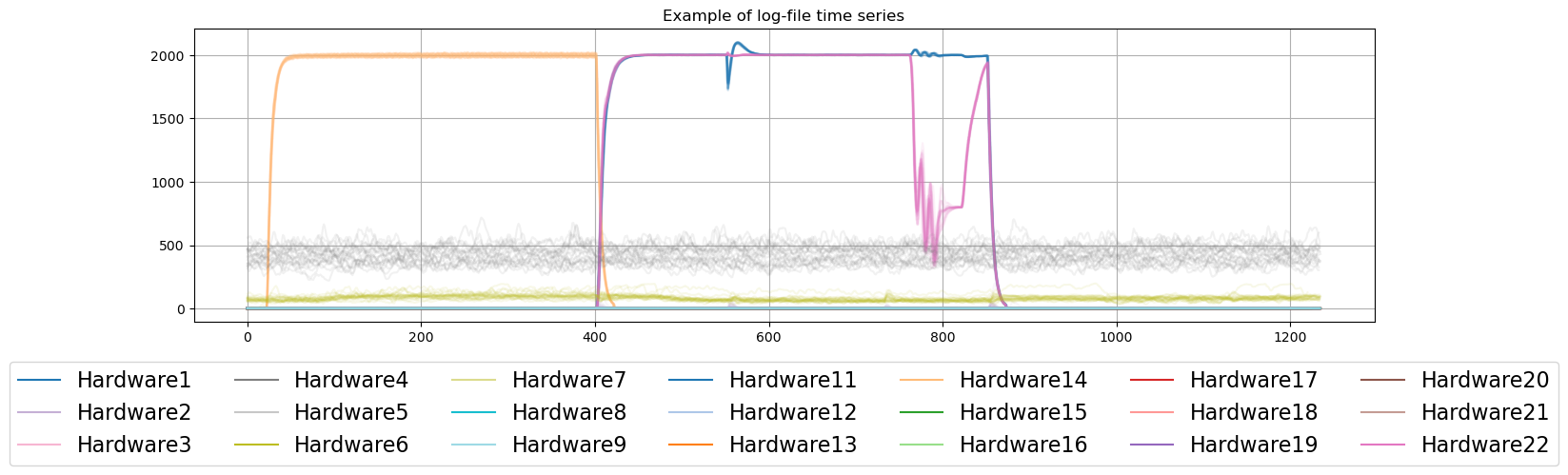}}
\caption{Example of a time series from a process in one chamber.}
\label{Fig:TimeSeries}
\end{figure*}

In this study, real-world data were collected from a Coat/Develop Track tool deployed in the \textit{imec} fabrication facility. In compliance with data confidentiality requirements, the variables have been anonymized and are presented in the generalized format \textbf{\textit{HardwareXX/VariableYY}}. The collected dataset comprises the multi-variate time series of 912 process runs from 14 distinct recipes and various chambers within the tool, sampled at 0.1-second time intervals. Since no labels indicating anomalous behaviour were provided, all data were assumed to be non-anomalous for the purpose of establishing proof-of-concept. An example of the multi-variate time series from a single process run is shown in Fig. \ref{Fig:TimeSeries}.

\subsection{Data preparation}\label{AA}
\subsubsection{Time series processing}
\begin{figure}[htbp]
\centerline{\includegraphics[width=0.5\textwidth]{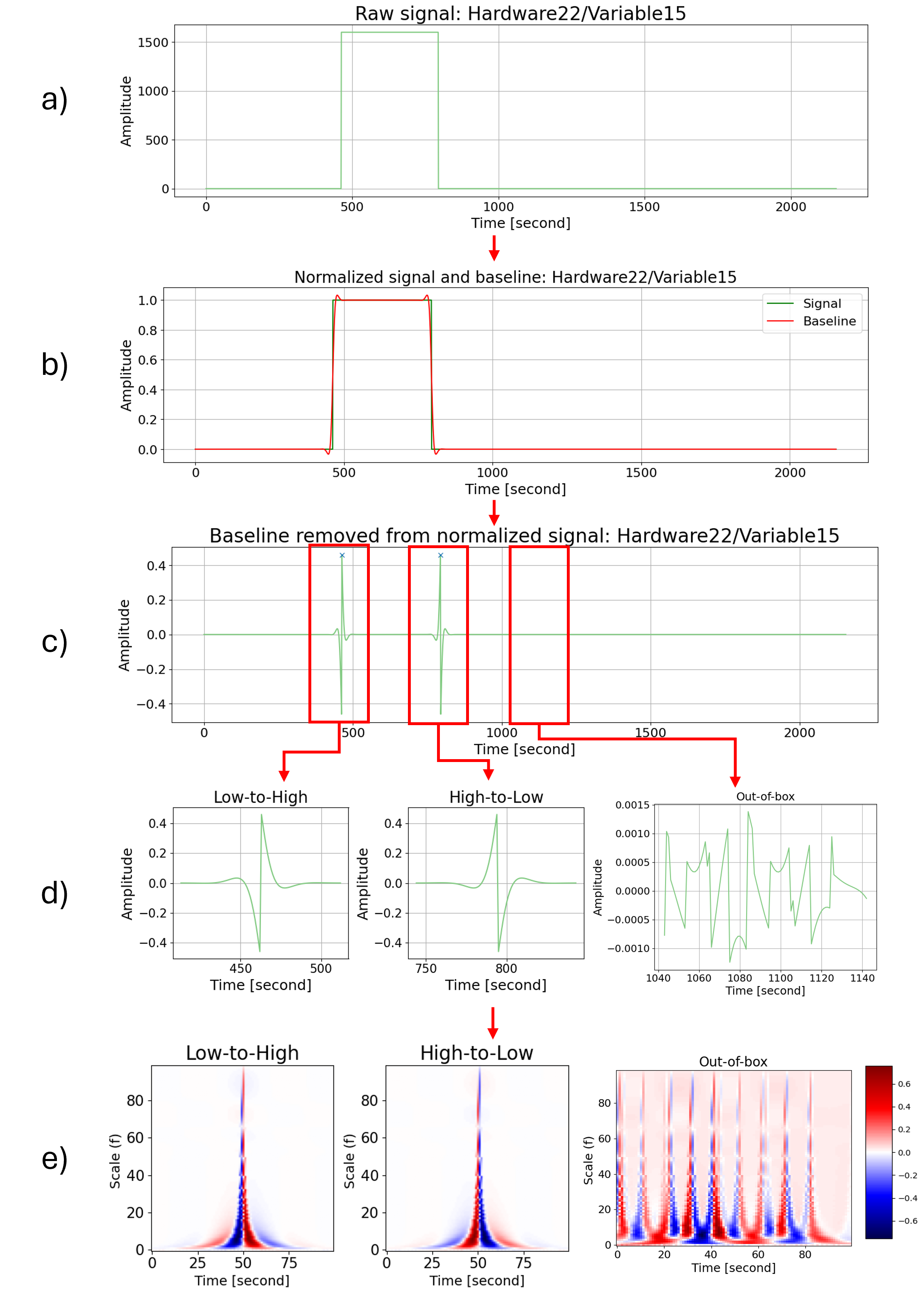}}
\caption{Pre-processing pipeline for converting time series data into Continuous Wavelet Transform (CWT) images. a) Raw sensor data; b) Normalized signal (green) with estimated baseline (red); c) Baseline-corrected signal; d) Zoom-in on two peak regions and an Out-of-Box (noisy) region; e) Corresponding CWT representations of the three regions shown in d).}
\label{fig:preprocessing}
\end{figure}

A common challenge in time series analysis is the high dimensionality of the data relative to the amount of relevant information it contains. As illustrated in Figure \ref{Fig:TimeSeries}, several of the collected time-series signals exhibit step-wise behaviour, a typical characteristic of variables whose values and timings are defined by the process recipe. Consequently, the relevant information can be effectively reduced to the timings and amplitude of these steps. To extract and process this information, the method illustrated in Fig. \ref{fig:preprocessing} is proposed.

The first step involves normalizing all time series using min-max scaling to ensure that the values lie within the interval [0,1]. Next, the baseline of each signal is estimated using asymmetric least squares \cite{Oller-Moreno2014}. This method is similar to ordinary least squares fitting, with an additional constraints to control the asymmetry and smoothness of the baseline. The baseline is computed by minimizing the cost function defined in (\ref{Eq:ASL}).

\begin{equation}
\begin{split}
    F = &\sum_i w_i\left(y_i - z_i\right)^2 \\
      &+ \lambda \sum_i \left(\Delta^2 z_i\right)^2
\end{split}
\label{Eq:ASL}
\end{equation}

Where $y_i$ denotes the original signal, $z_i$ the estimated baseline, $\Delta^2 z_i := z_i - 2z_{i-1} + z_{i-2}$, $w_i$ controls the asymmetry in the baseline estimation, while $\lambda$ regulates the degree of smoothness imposed on the estimated baseline. The first summation term ensures fidelity of the baseline to the original signal, whereas the second term enforces smoothness to $z$. In this work, $w_i$ was fixed at 0.5, effectively eliminating asymmetry from the baseline estimation.

This estimated baseline is subtracted from the original signal, isolating the step signatures -- sharp transitions occurring at step points, with amplitudes proportional to the magnitude of the steps. These signatures are detected by identifying local maxima with a minimum height of 0.1 and a minimum spacing of 10 seconds between peaks. A 10-second window centered around each peak is then extracted and converted into a spectrogram using the Continuous Wavelet Transform (CWT) \cite{Torrence1998}. Unlike the Fourier Transform \cite{Brigham1967}, which utilizes sine and cosine basis functions, the CWT employs wavelets. In this work, the Mexican hat wavelet was used \cite{Ryan1994}. The CWT is defined in (\ref{Eq:CWT}).

\begin{equation}
    \label{Eq:CWT}
    CWT_n(f) = \sum_{n'=0}^{N-1} x_{n'}\Psi^*\left[\frac{(n'-n)\delta t}{f}\right]
\end{equation}

where $x_n$ denotes the discrete time series of length $N$, with $\delta t = 0.1$ seconds representing the time step between consecutive values. $\Psi$ is the wavelet function used with $\Psi^*$ its complex conjugate, and $f$ denotes the scale parameter, which is analogous to frequency in the Fourier Transform. By applying this transformation, the signal is converted into a time-scale representation, effectively forming an image that captures the signal’s amplitude across different time points and scales. As illustrated in Fig. \ref{fig:preprocessing}.e), three distinct classes of peaks were identified: (1) Low-to-High transitions (blue on the left, red on the right), (2) High-to-Low transitions (red on the left, blue on the right) and (3) Out-of-Box (noisy signal). These distinct peaks can be used as input features for classification using well established image analysis models.

\subsubsection{Anomaly induction}
\begin{figure}[htbp]
\centering
\begin{subfigure}{0.47\textwidth}
    \centering
    \includegraphics[width=\textwidth]{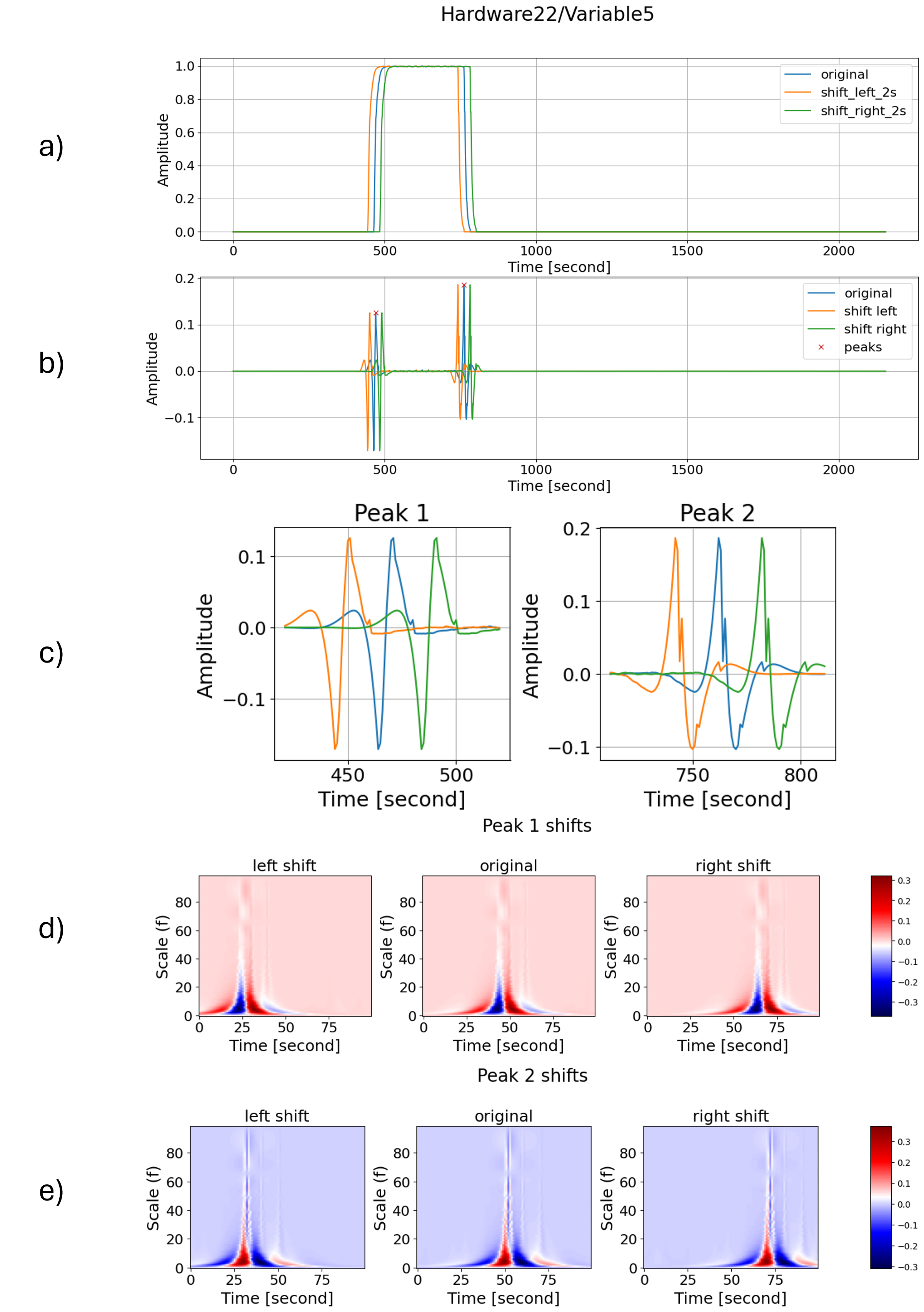}
    \caption{Induced time-shift anomaly and corresponding processing steps. a) Raw normalized sensor data and anomaly signals with $\pm$2-second time shifts; b) Corresponding baseline-corrected signals; c) Zoom-in on peak regions highlighting the $\pm$2-second shift of the peaks; d) CWT representations of the first peak in the 3 signals; e) CWT representations of the second peak in the 3 signals.}
    \label{fig:shifted}
\end{subfigure}
\hfill
\begin{subfigure}{0.51\textwidth}
    \centering
    \includegraphics[width=\textwidth]{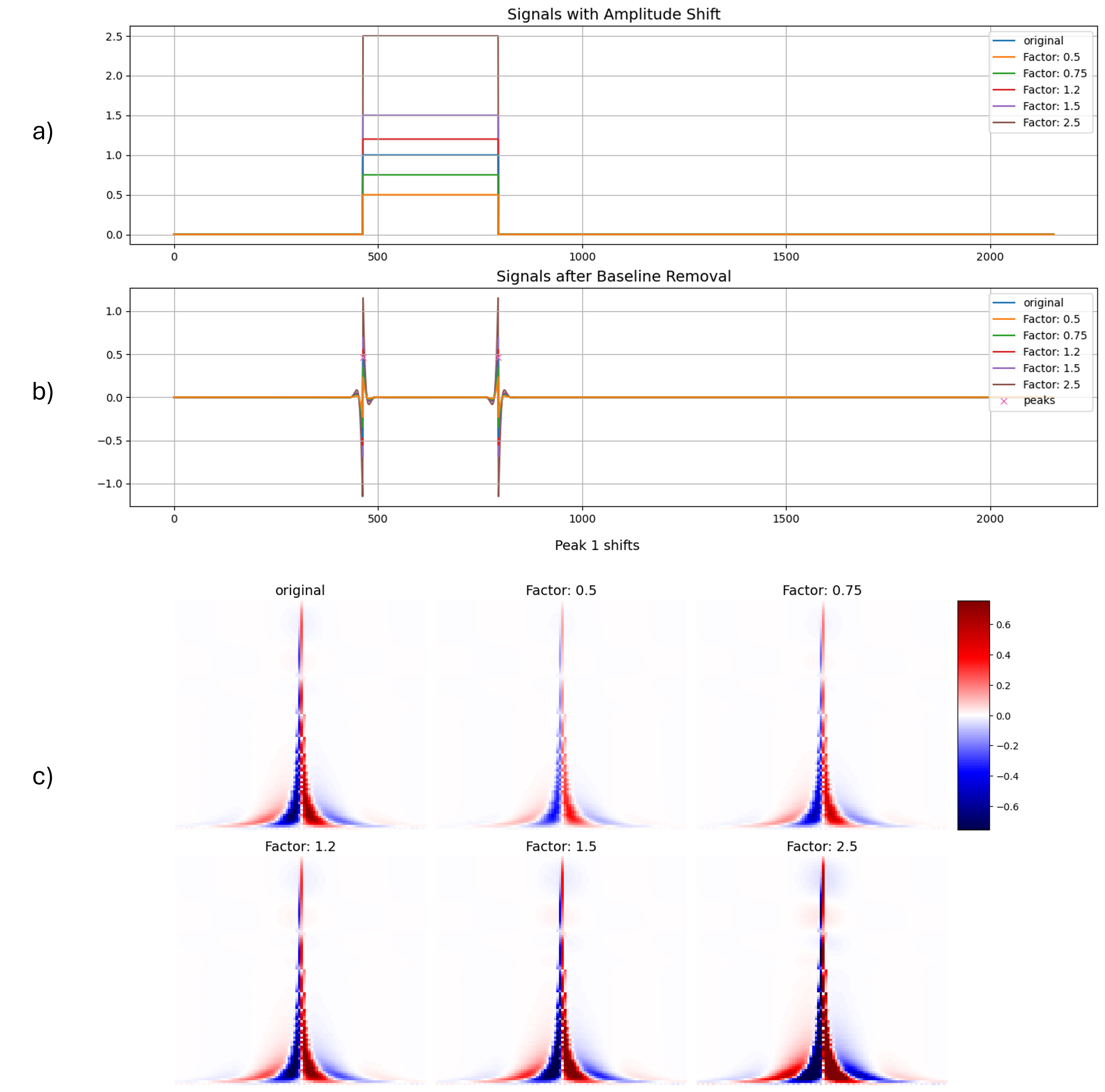}
    \vspace{1cm}
    \caption{Induced amplitude shift anomaly and corresponding processing steps. a) Raw normalized sensor data and anomaly signals; b) Corresponding baseline-corrected signals; c) CWT representations of the first peak in the 6 signals.}
    \vspace{1cm}
    \label{fig:AmpShift}
\end{subfigure}
\caption{Induced anomalies to form Dataset-2 and Dataset-3.}
\label{fig:Anomalies}
\end{figure}

As previously mentioned, the available dataset does not contain labelled anomalous events. Therefore, all collected data are assumed to represent non-anomalous instances, and anomalies must be artificially introduced. Among the most common deviations observed in trace data are \textbf{time shifts} and \textbf{amplitude shifts}, where signal steps occur earlier or later than expected, or exhibit altered amplitudes. These two anomaly categories were studied separately. \textbf{Time shifts} were induced by shifting the entire signal forwards or backwards in time. An example illustrating a 2-second time shift applied to the complete signal, along with subsequent processing steps is shown in Fig. \ref{fig:shifted}. \textbf{Amplitude shifts} were introduced by multiplying the step segments of the normal signal with constant factors. As is shown in Fig. \ref{fig:AmpShift}, the factors \{0.5, 0.75, 1.2, 1.5, 2.5\} were applied, resulting in signals with varying step amplitudes and corresponding CWT images exhibiting different colour intensities.

Using the generated anomalies, three datasets were constructed, the original dataset without anomalies, referred to as Dataset-1, an augmented dataset with time shift induced anomalies, referred to as Dataset-2, and a separate dataset with only amplitude shift-induced anomalies, referred to as Dataset-3.  Dataset-1 includes 56 images for each of the three classes (high-to-low: H\_L, low-to-high: L\_H, and out-of-box: O\_o\_B), resulting in a total of 168 images.  Dataset-2 extends Dataset-1 by adding 4 additional classes representing time-shifted anomalies (left and right shifts for for both H$\rightarrow$ L and L$\rightarrow$H, denoted as L\_H\_L, L\_H\_R, H\_L\_L and H\_L\_R) with 56 images per new class. This results in a total of 7 classes and 392 images. Finally, Dataset-3 consists of a total of 12 CWT images: 2 images per peak from the original (non-anomalous) signal, and the remaining 10 images (5 per peak) generated by applying amplitude shift factors to the original signal, as shown in Fig. \ref{fig:AmpShift}. It is important to note that, for this dataset, only one image is available for each instance of the 'normal/non-anomalous' and 'anomalous' classes. This presents a fundamental challenge, as the model (described in section \ref{subsec:Model}) is likely to overfit on Dataset-3 due to the limited data. Although overfitting is typically undesirable, Dataset-3 serves as a controlled test case to evaluate whether the proposed methods can still effectively detect and distinguish amplitude deviations.


\begin{figure*}[h]
    \centering
    \begin{subfigure}[b]{\textwidth}
        \centering
        \includegraphics[width=\textwidth]{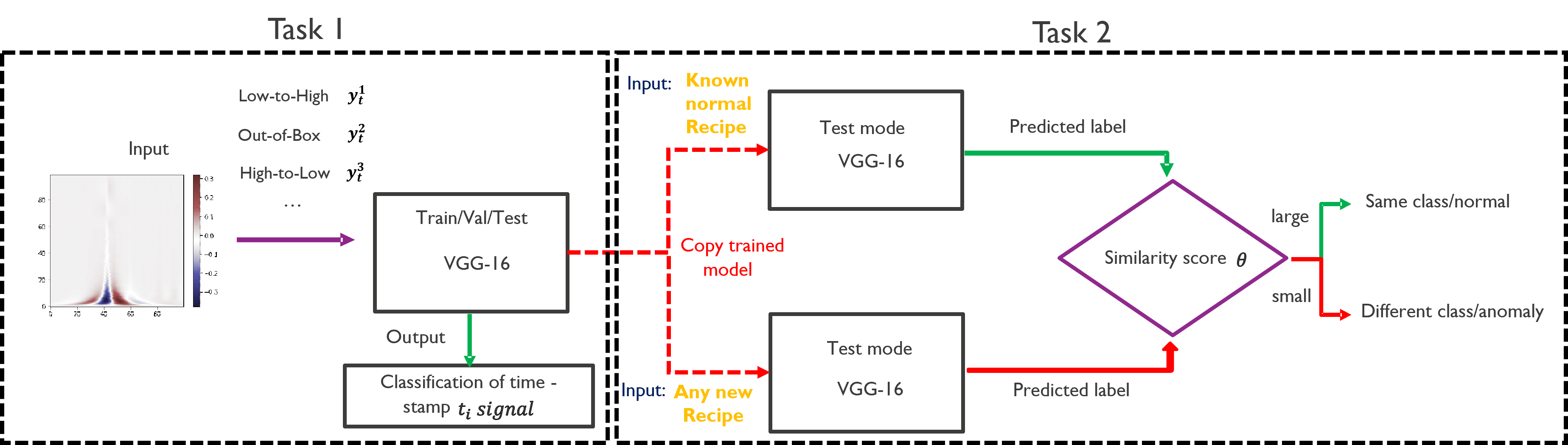}
        \caption{Workflow}
        \label{subfig:workflow}
    \end{subfigure}
    \vspace{1cm}
    \begin{subfigure}[b]{0.4\textwidth}
        \centering
        \includegraphics[width=\textwidth, height=5cm]{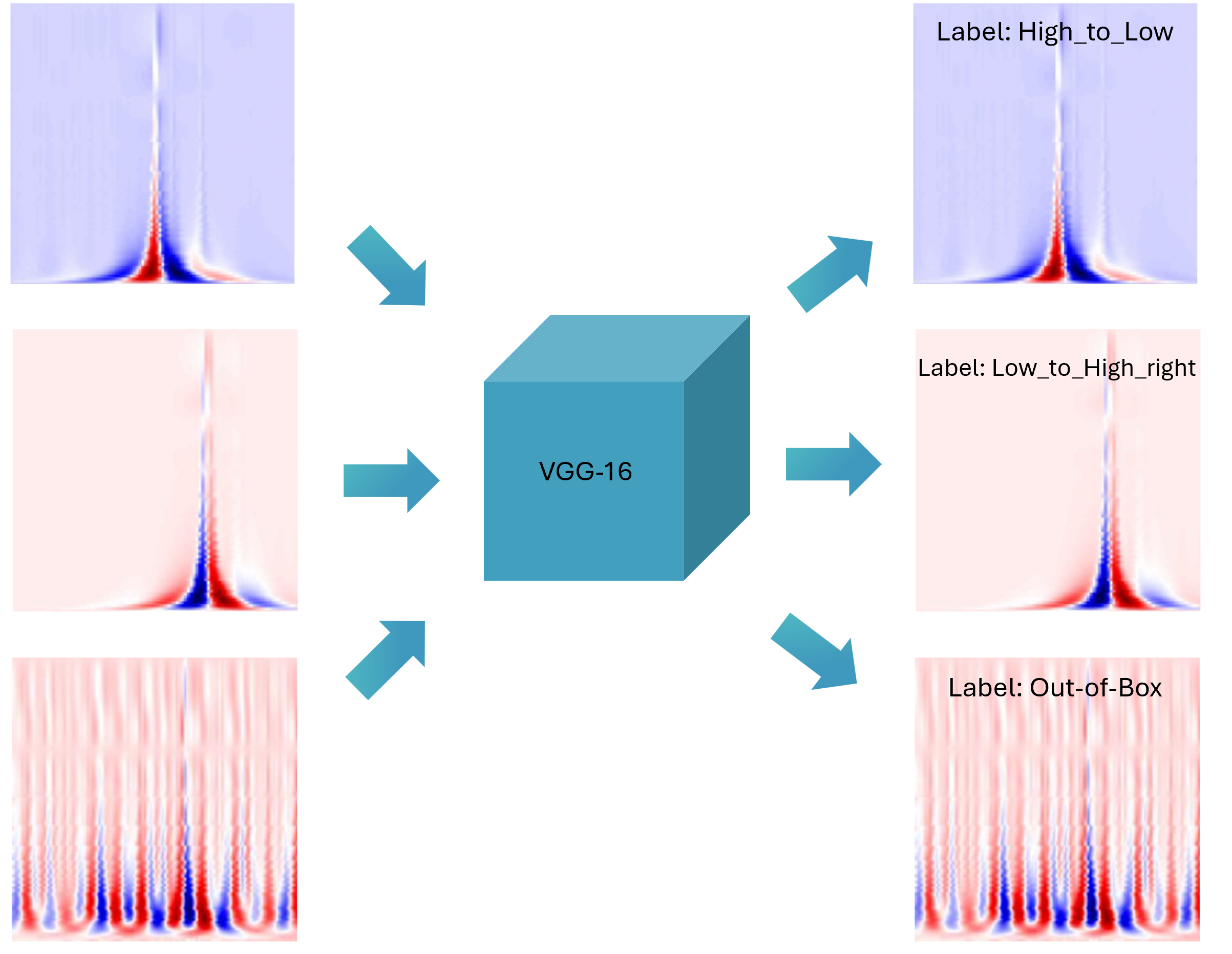}
        \caption{Task 1}
        \label{subfig:step1}
    \end{subfigure}
    \hfill
    \begin{subfigure}[b]{0.5\textwidth}
        \centering
        \includegraphics[width=\textwidth, height=5cm]{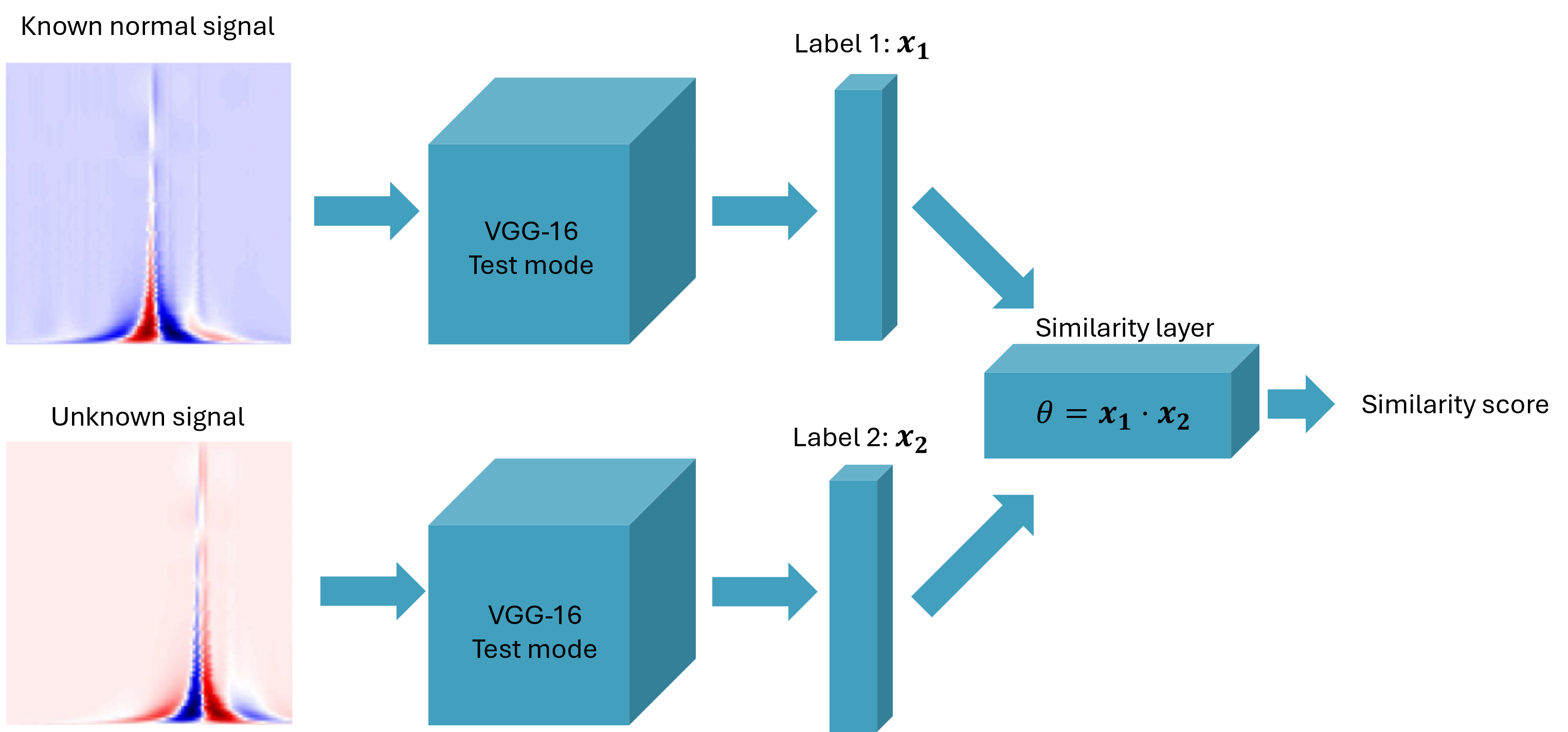}
        \caption{Task 2}
        \label{subfig:step2}
    \end{subfigure}
    \caption{Proposed Anomaly Detection and Localization Framework. (a) Overview of the complete framework, divided into two primary tasks. (b) Task 1: Train a CNN architecture to classify each CWT input image (x$^i_{t_j}$) with its corresponding process state label (y$^i_{t_j}$). (c) Task 2: Employ a Siamese network -- comprising two identical sub‑networks that share the fine‑tuned CNN backbone -- to compare pairs of CWT images (x$^i_{t_j}$, x'$^i_{t_j}$). A similarity score between their predicted labels determines whether the new process/tool trace exhibits normal or anomalous behaviour.}
    \label{fig:workflow}
\end{figure*}

\subsection{Model and training structure}
\label{subsec:Model}
The backbone of the model employed in this work is the \textit{VGG-16} architecture \cite{Simonyan15}, a deep Convolutional Neural Network developed for image analysis and classification tasks. Pre-trained weights from the \textit{ImageNet} \cite{ILSVRC15} dataset are utilized, with the final four layers of the network frozen to retain high-level feature extraction capabilities. This approach leverages transfer learning to minimize training time and enhance the model's adaptability to our target dataset.

The backbone model was applied to two sequential tasks on all datasets: 1. \textbf{Image classification task}: Each CWT image x$^i$ at time-step $t_j$ is classified into its corresponding state class. The classifier trained in this task must learn the CWT signatures associated with different process states. For example, amplitude variations are reflected as localized intensity changes (i.e., brighter or darker regions) at specific scales and time points, while time-drift anomalies appears as horizontal shifts of similar patterns across the time axis; 2. \textbf{Siamese similarity scoring}: The fine‑tuned classifier is repurposed within a Siamese network to compare pairs of images (x$^i_{t_j}$, x'$^i_{t_j}$) at the same time-step $t_j$. The Siamese architecture consists of two identical branches, both initialized with the trained classifier. One branch receives a reference non‑anomalous image x$^i_{normal}$ at $t_j$ , while the other processes a new image x$^i_{unknown}$ at $t_j$. A high similarity score signifies normal behaviour; a low score indicates an anomaly. The full training pipeline and operational workflow are depicted in Fig. \ref{fig:workflow}.

For Task-1, additional pooling and dense layers were appended to the VGG-16 model, ensuring the output shape matched the number of target classes, followed by a softmax activation. The dataset was divided into 70\% for training and 30\% for testing. To address the limited sample size, data augmentation was applied during training, including rotations, horizontal and vertical flips, contrast adjustments, and pixel dropout, altering approximately half of the 32 images in each batch. This strategy increased input diversity and helped reduce overfitting. Model performance was assessed using a confusion matrix computed from the test set.

For Task-2, the classifier trained in Task 1 is repurposed within a Siamese architecture: two identical model copies operate in parallel, each processing one CWT image (x$^i_{t_j}$ and x'$^i_{t_j}$) at time-step $t_j$. Each branch outputs a probability vector indicating the likelihood that its CWT image belongs to each class. To measure similarity, we compute the dot product of these two vectors: values near 1.0 imply the images share the same class, whereas values near 0.0 indicate different classes. In practice, one branch processes a reference image from a known non‑anomalous run, and the other evaluates a new image. If no anomaly has occurred, both images map to the same class and yield a high similarity score. Conversely, an anomaly alters the process behavior, causing the CWT images to fall into different classes and produce a low similarity score.

The dataset for this task was partitioned into 75\% for training and 25\% for testing. During training, data augmentation as contrast enhancement via histogram equalization was applied for each of the three channels (RGB). This augmentation was particularly beneficial for the second dataset, where time-shift anomalies needed to be identified and higher contrast aids in detecting these shifts. For the evaluation of this task, N-way validation was computed. This consists in sampling an anchor image from the test data as well as \textit{N} other images out of which one belongs to the same class as the anchor image. The model is then tested by computing the similarity score between the anchor image and the \textit{N} other images so the one with the highest similarity score is predicted to be of the same class. By performing this test \textit{k}-times, the fraction of times the model correctly predicts which image belonged to the same class as the anchor image is the metric used to evaluate the performance of the model. The value of \textit{k} should be high enough for every image in the test dataset to be picked at least once. The expected value of \textit{k} is the solution to the Coupon collector's problem \cite{Erds1961} and is easily calculated as $k=\sum_{i=1}^M \frac{M}{i}$, where $M$ denotes the number of images in the test dataset.

\section{Results \& Discussion}
\label{Sec:R_D}

As described in Section \ref{Sec:Methodology}, three datasets were created: Dataset-1,  containing 3 non-anomalous classes; Dataset-2, which extends Dataset-1 by adding 4 anomaly classes; and Dataset-3, which comprises 12 classes-2 non-anomalous and 10 anomalous. As each dataset was designed to serve a specific objective, their respective results will be discussed separately in this section.

\subsection{Dataset-1: Non-Anomalous}

\begin{figure}[htbp]
\centering
\begin{subfigure}{0.45\textwidth}
    \includegraphics[width=\textwidth]{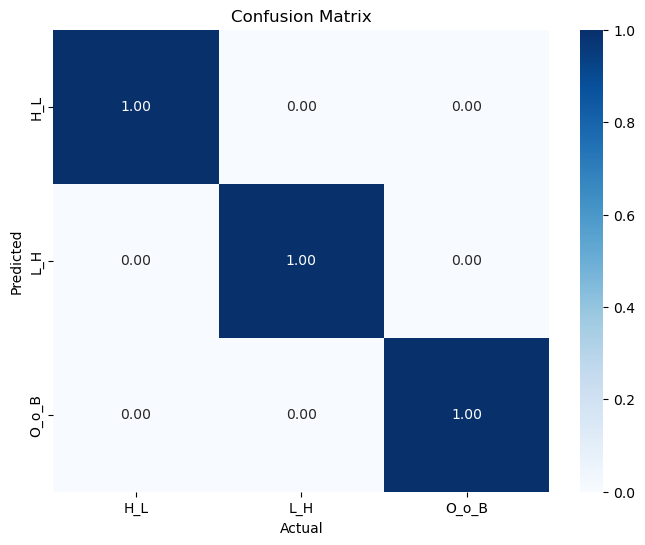}
    \caption{Confusion matrix for classifier model on Dataset-1}
    \label{fig:D1-T1}
\end{subfigure}
\hfill
\begin{subfigure}{0.45\textwidth}
    \includegraphics[width=\textwidth]{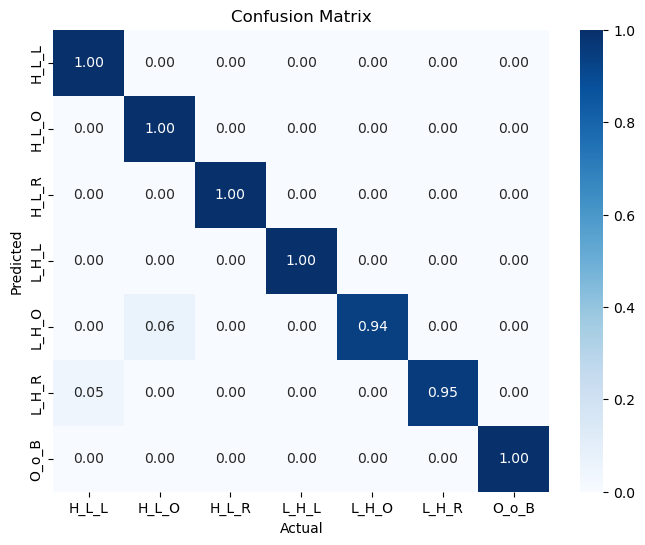}
    \caption{Confusion matrix for classifier model on Dataset-2}
    \label{fig:D2-T1}
\end{subfigure}
\caption{Confusion matrices for classifier models on Dataset-1 and Dataset-2}
\label{fig:combined}
\end{figure}

For the image classification task (Task-1) on Dataset-1, the classifier was trained to learn the CWT signatures of non-anomalous signal steps (i.e., known normal signal peaks) as well as idle periods (labelled as O\_o\_B). The confusion matrix in Fig. \ref{fig:D1-T1} illustrates the prediction accuracy achieved on the test set. The classifier successfully classified all images correctly, demonstrating its effectiveness in capturing the intended signal characteristics. When applied within the Siamese framework for Task-2, a 20-way validation with \textit{k} = 182 resulted in a perfect 100\% identification accuracy. This outcome is consistent with the classifier's optimal performance in Task-1.

\subsection{Dataset-2: Time Shifts}

In Dataset-2, the objective was to learn and differentiate time-shifted steps from known normal steps and idle periods. The performance for Task-1 is presented in the confusion matrix in Fig. \ref{fig:D2-T1}, which shows a near perfect average accuracy of 99\%, with only 2 misclassifications.

\begin{figure}[htbp]
\centerline{\includegraphics[width=0.45\textwidth]{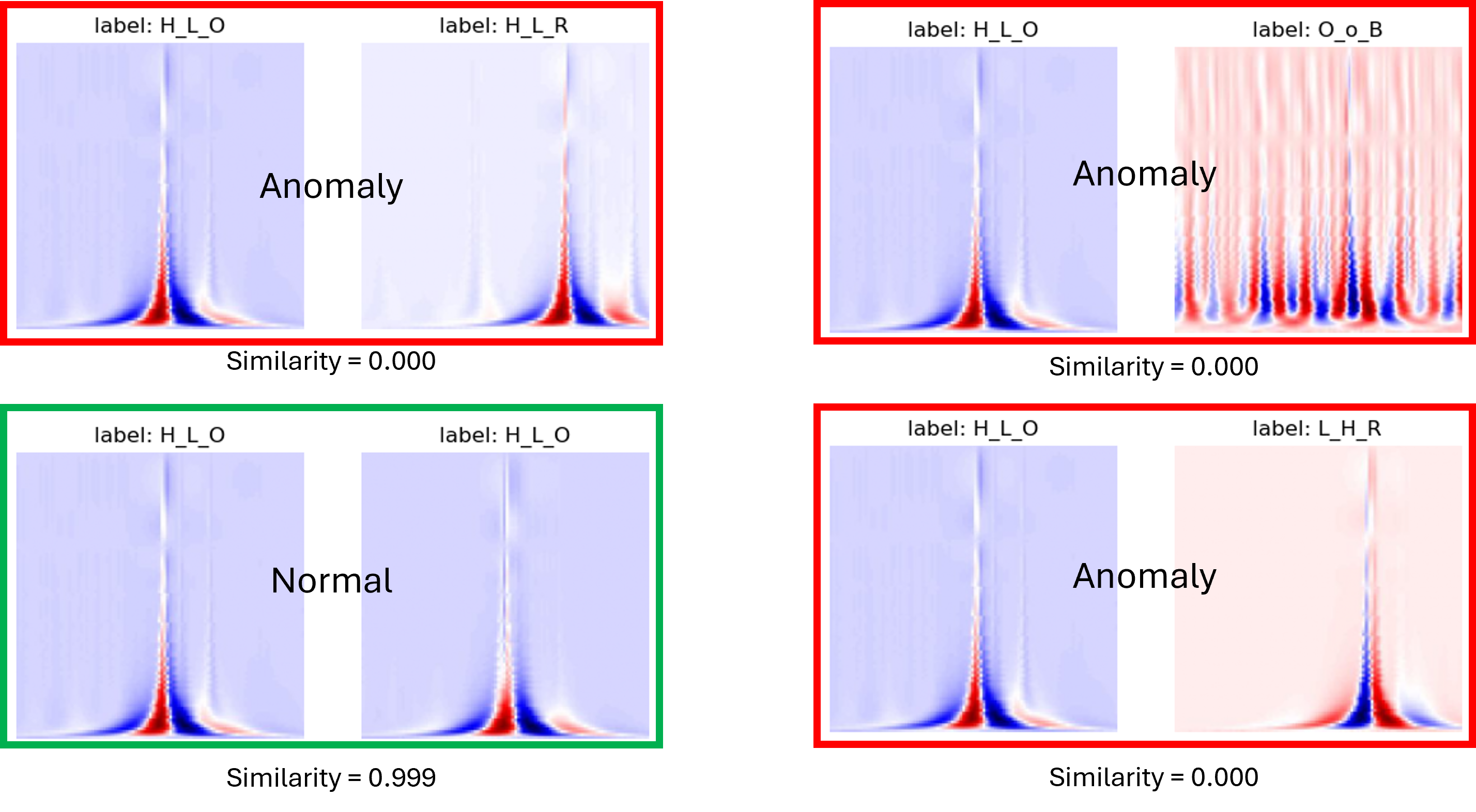}}
\caption{Siamese Input and output during N-way validation of Dataset-2. Samples an anchor image and $N$ other images with only one of the $N$ images belong to the same class as the anchor image. In the illustrated case 4-way validation is performed, with the left image of each pair being the anchor image.}
\label{fig:SiaResults}
\end{figure}

For Task-2, a 20-way validation with \textit{k} = 506 resulted in a perfect 100\% identification accuracy. Although a similar error rate is expected across both Tasks 1 and 2 for Dataset-2, the high value of $k< 100$ leads to a dilution of misclassifications, effectively rounding the overall accuracy up to 100\% in the N-way validation procedure. As an illustrative example, the Siamese model's prediction results for a single trial, 4-way validation are presented in Fig. \ref{fig:SiaResults}.

\subsection{Dataset-3: Amplitude Shifts}
\begin{figure}[htbp]
\centerline{\includegraphics[width=0.45\textwidth]{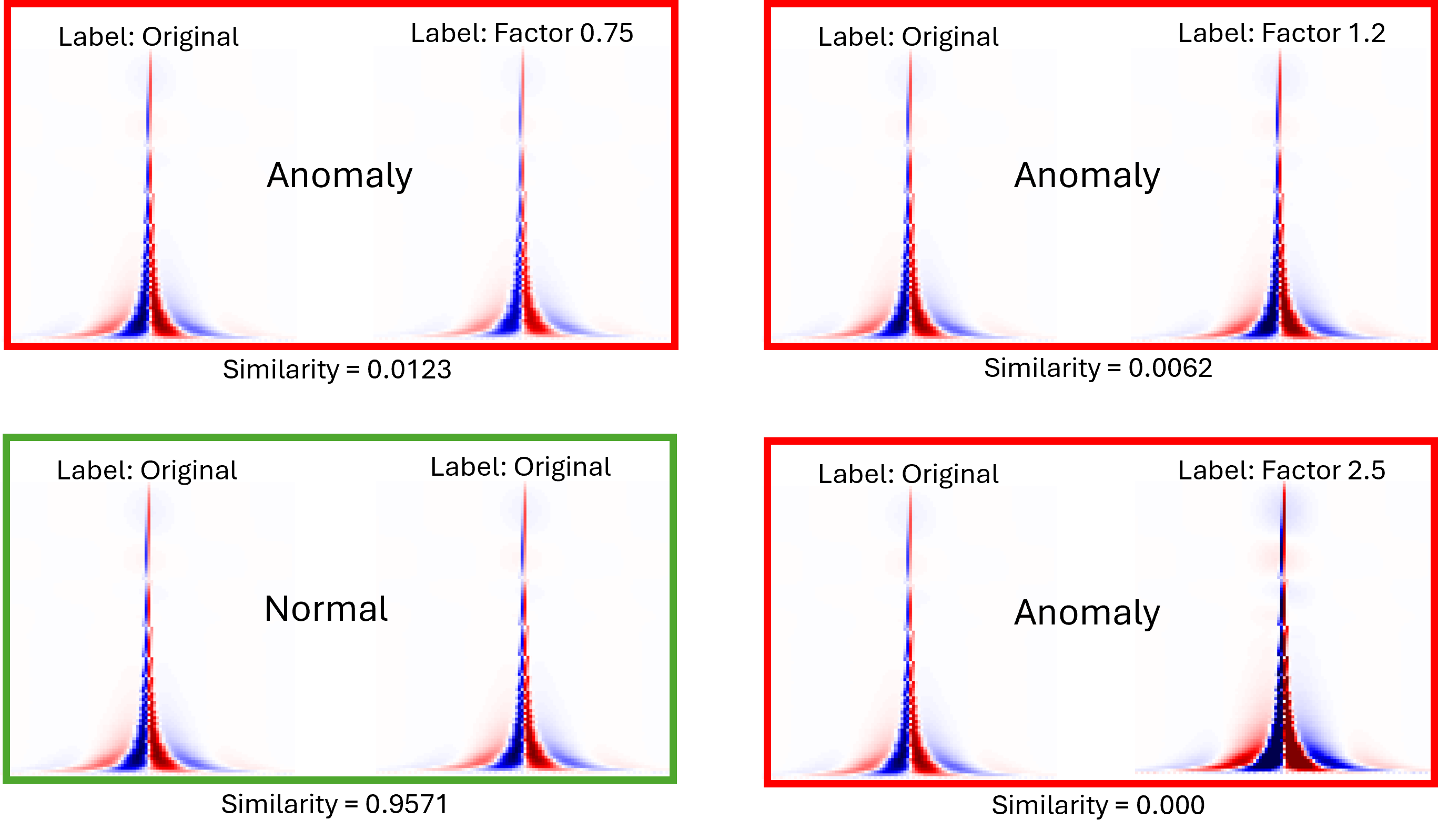}}
\caption{Siamese Input and Output for selected images from Dataset-3. The model clearly identifies the non-anomalous image by assigning a high similarity score, while similarity scores progressively decrease as the compared images exhibit increasing levels of amplitude deviation from the original.}
\label{fig:AmpResults}
\end{figure}

Lastly Dataset-3 was used to evaluate the proposed framework’s capability to detect amplitude shift-induced anomalies. Since this dataset contains only one CWT image per class, it is not feasible to compute classification accuracy on unseen data for Task-1. This is because the normal signal (peak amplitude) exhibits a distinct intensity profile, while any amplitude-shifted anomaly results in a different intensity distribution.Therefore, no confusion matrix is presented for this task. For Task-2, the 12 CWT images were divided into two groups based on the peak (either known-normal peak-1 or peak-2) from which the amplitude-shifted images were derived. Within each group, the image representing the original (unshifted) signal served as the “anchor" or “known-normal" reference and was compared-via the Siamese framework-to itself (factor = 1.0) and to the five anomalous (amplitude-shifted) images within the group. Representative examples of the Siamese framework's input and output are illustrated in Fig. \ref{fig:AmpResults}. Instead of N-way validation, a similarity score-based evaluation was performed, as presented in Table \ref{Tab:AmpShift}. As expected, the “anchor" image achieved the highest similarity with itself (factor = 1.0), with observed similarity values of $\sim 0.9571$ for peak-1 and $\sim 0.9257$ for peak-2. The similarity scores progressively decreased as the compared images exhibited increasing levels of amplitude deviation from the original. These results confirm that the proposed framework effectively detects amplitude shift-induced anomalies, ranging from subtle to pronounced deviations, and that the similarity (or dissimilarity) score reliably captures and quantifies the degree of deviation from the original signal.

To further improve the model’s robustness, future work
should focus on improving both sensitivity and generalizability
by increasing the diversity and scale of the training data,
particularly to capture subtle variations in process behaviour, especially for Task-2 (in Dataset-2 and Dataset-3), building upon the foundation established in Task-1.

\begin{table}[htpb]
\centering
\resizebox{0.8\textwidth}{!}{
\begin{tabular}{|c|c|c|c|c|c|c|}
\cline{1-3} \cline{5-7}
\multirow{7}{*}{Peak 1} & Amplitude Shift Factor & Similarity Score &  & \multirow{7}{*}{Peak 2}  & Amplitude Shift Factor & Similarity Score \\\cline{2-3} \cline{6-7}
                        & Factor: 0.5            & 0.0001           &   &                         & Factor: 0.5            & 0.0                \\
                        & Factor: 0.75           & 0.0123           &   &                         & Factor: 0.75           & 0.0005           \\
                        & \textbf{Factor: 1}              & \textbf{0.9571}           &   &                         & \textbf{Factor: 1}              & \textbf{0.9257}           \\
                        & Factor: 1.2            & 0.0062           &   &                         & Factor: 1.2            & 0.0343           \\
                        & Factor: 1.5            & 0.0                &   &                         & Factor: 1.5            & 0.0                \\
                        & Factor: 2.5            & 0.0001           &   &                         & Factor: 2.5            & 0.0                \\
\cline{1-3} \cline{5-7}
\end{tabular}}
\vspace{0.2cm}
\caption{Similarity scores for amplitude-shifted images in Dataset-3. Dataset was split into 2 groups, one for each peak, and in each group the similarity between the original image and each amplitude-shifted image was computed.}
\label{Tab:AmpShift}
\end{table}

\section{Limitations and Future Work}
\label{Sec:Limitations}
 
 While our methodology integrates time–frequency analysis, deep spatial feature extraction, and metric-based comparison to address many of the limitations inherent in classical statistical and conventional machine learning approaches, it presents several limitations that can be addressed in future work:
 \subsection{Data requirements and generalization:} The proposed approach relies on a representative set of known normal (non-anomalous) reference traces. Due to tool data privacy constraints, the available raw time-series data were assumed to be non-anomalous. To enable anomaly detection evaluation, synthetic anomalies, validated through scientific and mathematical reasoning, were introduced in the form of time shifts and amplitude variations. However, limited coverage of the full variability in normal operating conditions may affect the model’s generalizability. Additionally, rare or previously unseen anomalies that are not reflected in the reference library may go undetected. Domain shifts, such as modifications in tool hardware, process recipes, or sensor configurations, can significantly alter the CWT signature of normal behavior, potentially requiring re-collection of reference traces or model retraining. Future research should explore self-supervised learning strategies to mitigate these limitations, and validate the model's generalizability against real-world anomalous instances across a broad range of tools and process conditions.
\subsection{Computational and Memory overhead:} The proposed approach imposes considerable computational and memory demands, which may impede real-time operation and edge deployment. This motivates further exploration of optimized CNN variants, through techniques like pruning and quantization, and to develop lightweight CWT approximations to facilitate practical implementation on fab tools. Additionally, embedding active or continual learning within a federated framework could automate the updating of reference libraries and model parameters, minimizing human intervention and improving adaptability to frequent changes in fab conditions.

\subsection{Hyperparameter sensitivity:} The selection of wavelet, scale range, window size, and overlap critically influences performance and currently relies on manual expert tuning. Suboptimal configurations can mask critical anomalies, and finding the right trade-off between temporal resolution (smaller windows) and frequency resolution (larger windows) demands systematic study. Additionally, determining a suitable distance threshold for anomaly detection, especially in unsupervised contexts, is non-trivial and often necessitates periodic revalidation. Future research should aim to automate these selections via Bayesian optimization or meta-learning, incorporate adaptive thresholding mechanisms, and leverage neural architecture search (NAS) \cite{elsken2019neuralarchitecturesearchsurvey} to identify optimal wavelet parameters.

\subsection{Temporal context limitations:} Our current framework evaluates each time step independently, using only the information contained within a single, fixed-length window. Consequently, it may overlook slowly evolving patterns, such as gradual sensor drifts or subtle cyclical behaviors that span multiple windows, and independent frame scoring can produce inconsistent labels over time. Future strategy to integrate CNN model with sequential architectures (e.g., RNNs\cite{schmidt2019recurrentneuralnetworksrnns} or Transformers\cite{Khan_2022}) or multi-scale aggregation methods, as well as incorporating temporal smoothing techniques (such as Conditional Random Fields\cite{10.5555/645530.655813} to enforce label consistency across adjacent frames. 

\subsection{Interpretability constraints:} Finally, although our proposed method can highlight salient time-frequency regions, interpreting these complex patterns is challenging and still requires domain expertise. Moreover, a high distance score indicates only a deviation, but does not specify the anomaly's precise nature (e.g., amplitude variation vs. temporal drift) without additional analysis. Future work will focus on integrating explainable-AI modules (such as attention maps\cite{10780702}, prototype-based explanations) directly within Siamese CNN's embedding space. Additionally, we plan to develop an interactive visualization user-interface (UI) that links highlighted spectrogram regions to their corresponding physical process or tool parameters.

\section{Conclusion}
\label{Sec:Conclusion}
In this study, we introduce a novel deep-learning framework designed to address the challenges of multivariate time-series processing and anomaly detection in semiconductor process and tool traces. First, we employ the CWT to preprocess raw signals into fixed-size time-frequency images, effectively capturing both transient spikes and gradual drifts. Next, a Siamese-CNN compares each incoming CWT image to a library of known-good references, enabling robust, fine-grained, and scalable detection of drifts and anomalies. Our proposed approach outperforms traditional statistical control schemes and many supervised ML models, particularly in handling transient spikes, oscillatory patterns, non-linear dynamics, and the high-dimensional data typical of fab processes. Unlike fixed-threshold approaches that must be manually recalibrated for each tool, chamber, or recipe, our metric-learning strategy automatically adapts to evolving “normal” patterns, offering robust resilience against process drift and tool variability. By highlighting the specific time-frequency regions responsible for increased distance scores, the framework also enhances explainability and aids engineers in root-cause analysis. Finally, because it trains primarily on abundant normal data, with only a few labeled anomalies when available, it coherently supports both supervised and semi-supervised deployment, making it a practical solution for detecting rare or unforeseen faults in advanced process control.

\section{Acknowledgement}
\label{Sec:Acknowledgement}

The authors would like to express their sincere gratitude to \textbf{Dr. Yasutoshi Okuno}, \textbf{Mr. Jun Kawai}, and \textbf{Mr. Hiroshi Horiguchi} from \textbf{SCREEN Holdings Co., Ltd.}, \textbf{SCREEN Advanced System Solutions Co., Ltd.}, and \textbf{SCREEN Semiconductor Solutions Co., Ltd.} for their valuable insights, constructive feedback, and engaging discussions, which significantly contributed to the direction, execution, and interpretation of this research work.



\begin{thebibliography}{10}

\bibitem{FlowErrorCosts}
Error in measuring low flows could cost chipmakers millions | machine design.
\newblock URL: \url{https://www.machinedesign.com/mechanical-motion-systems/article/21837100/error-in-measuring-low-flows-could-cost-chipmakers-millions}.

\bibitem{DowntimeCosts}
Getting smarter about tool maintenance.
\newblock URL: \url{https://semiengineering.com/getting-smarter-about-tool-maintenance/}.

\bibitem{Lapedes1987}
A~Lapedes and R~Farber.
\newblock Nonlinear signal processing using neural networks: Prediction and system modelling, 6 1987.

\bibitem{ARIMA}
D.~J. Bartholomew.
\newblock Time series analysis forecasting and control.
\newblock {\em Journal of the Operational Research Society}, 22:199--201, 6 1971.
\newblock URL: \url{https://www.tandfonline.com/doi/abs/10.1057/jors.1971.52}, \href {https://doi.org/10.1057/JORS.1971.52} {\path{doi:10.1057/JORS.1971.52}}.

\bibitem{ARCH}
Robert~F. Engle.
\newblock Autoregressive conditional heteroscedasticity with estimates of the variance of united kingdom inflation.
\newblock {\em Econometrica}, 50:987, 7 1982.
\newblock \href {https://doi.org/10.2307/1912773} {\path{doi:10.2307/1912773}}.

\bibitem{MLP}
Alfonso Palmer, Juan~José Montaño, and Albert Sesé.
\newblock Designing an artificial neural network for forecasting tourism time series.
\newblock {\em Tourism Management}, 27:781--790, 10 2006.
\newblock URL: \url{https://www.sciencedirect.com/science/article/abs/pii/S0261517705000555}, \href {https://doi.org/10.1016/J.TOURMAN.2005.05.006} {\path{doi:10.1016/J.TOURMAN.2005.05.006}}.

\bibitem{TCN}
Shaojie Bai, J.~Zico Kolter, and Vladlen Koltun.
\newblock An empirical evaluation of generic convolutional and recurrent networks for sequence modeling.
\newblock 3 2018.
\newblock URL: \url{https://arxiv.org/pdf/1803.01271}.

\bibitem{LSTM}
Thomas Fischer and Christopher Krauss.
\newblock Deep learning with long short-term memory networks for financial market predictions.
\newblock {\em European Journal of Operational Research}, 270:654--669, 10 2018.
\newblock URL: \url{https://www.sciencedirect.com/science/article/pii/S0377221717310652}, \href {https://doi.org/10.1016/J.EJOR.2017.11.054} {\path{doi:10.1016/J.EJOR.2017.11.054}}.

\bibitem{SAEexample}
Pengfei Zhang, Xiaoping Ma, Wenyu Zhang, Shaowei Lin, Huilin Chen, Arthur~Lee Yirun, and Gaoxi Xiao.
\newblock Multimodal fusion for sensor data using stacked autoencoders.
\newblock {\em 2015 IEEE 10th International Conference on Intelligent Sensors, Sensor Networks and Information Processing, ISSNIP 2015}, 5 2015.
\newblock \href {https://doi.org/10.1109/ISSNIP.2015.7106972} {\path{doi:10.1109/ISSNIP.2015.7106972}}.

\bibitem{Torrence1998}
Christopher Torrence and Gilbert~P Compo.
\newblock A practical guide to wavelet analysis.
\newblock {\em Bulletin of the American Meteorological Society}, 79:61 -- 78, 1998.
\newblock URL: \url{https://journals.ametsoc.org/view/journals/bams/79/1/1520-0477_1998_079_0061_apgtwa_2_0_co_2.xml}, \href {https://doi.org/10.1175/1520-0477(1998)079<0061:APGTWA>2.0.CO;2} {\path{doi:10.1175/1520-0477(1998)079<0061:APGTWA>2.0.CO;2}}.

\bibitem{Brigham1967}
E.~O. Brigham and R.~E. Morrow.
\newblock The fast fourier transform.
\newblock {\em IEEE Spectrum}, 4:63--70, 1967.
\newblock \href {https://doi.org/10.1109/MSPEC.1967.5217220} {\path{doi:10.1109/MSPEC.1967.5217220}}.

\bibitem{ZamanzadehDarban2024}
Zahra~Zamanzadeh Darban, Geoffrey~I. Webb, Shirui Pan, Charu Aggarwal, and Mahsa Salehi.
\newblock Deep learning for time series anomaly detection: A survey.
\newblock {\em ACM Computing Surveys}, 57:42, 1 2024.
\newblock URL: \url{https://dl.acm.org/doi/10.1145/3691338}, \href {https://doi.org/10.1145/3691338/SUPPL_FILE/3691338.PDF} {\path{doi:10.1145/3691338/SUPPL_FILE/3691338.PDF}}.

\bibitem{Chen2020}
Chieh~Yu Chen, Shi~Chung Chang, and Da~Yin Liao.
\newblock Equipment anomaly detection for semiconductor manufacturing by exploiting unsupervised learning from sensory data.
\newblock {\em Sensors 2020, Vol. 20, Page 5650}, 20:5650, 10 2020.
\newblock URL: \url{https://www.mdpi.com/1424-8220/20/19/5650/htm https://www.mdpi.com/1424-8220/20/19/5650}, \href {https://doi.org/10.3390/S20195650} {\path{doi:10.3390/S20195650}}.

\bibitem{Mellah2022}
Samia Mellah, Youssef Trardi, Guillaume Graton, Bouchra Ananou, El~Mostafa~E.L. Adel, and Mustapha Ouladsine.
\newblock Semiconductor multivariate time-series anomaly classification based on machine learning ensemble techniques*.
\newblock {\em IFAC-PapersOnLine}, 55:476--481, 1 2022.
\newblock \href {https://doi.org/10.1016/J.IFACOL.2022.07.174} {\path{doi:10.1016/J.IFACOL.2022.07.174}}.

\bibitem{Baek2023}
Minjae Baek and Seoung~Bum Kim.
\newblock Failure detection and primary cause identification of multivariate time series data in semiconductor equipment.
\newblock {\em IEEE Access}, 11:54363--54372, 2023.
\newblock \href {https://doi.org/10.1109/ACCESS.2023.3281407} {\path{doi:10.1109/ACCESS.2023.3281407}}.

\bibitem{Hwang2023}
Rakhoon Hwang, Seungtae Park, Youngwook Bin, and Hyung~Ju Hwang.
\newblock Anomaly detection in time series data and its application to semiconductor manufacturing.
\newblock {\em IEEE Access}, 11:130483--130490, 2023.
\newblock \href {https://doi.org/10.1109/ACCESS.2023.3333247} {\path{doi:10.1109/ACCESS.2023.3333247}}.

\bibitem{Oller-Moreno2014}
Sergio Oller-Moreno, Antonio Pardo, Juan~Manuel Jimenez-Soto, Josep Samitier, and Santiago Marco.
\newblock Adaptive asymmetric least squares baseline estimation for analytical instruments.
\newblock In {\em 2014 IEEE 11th International Multi-Conference on Systems, Signals \& Devices (SSD14)}, pages 1--5. IEEE, 2 2014.
\newblock URL: \url{http://ieeexplore.ieee.org/document/6808837/}, \href {https://doi.org/10.1109/SSD.2014.6808837} {\path{doi:10.1109/SSD.2014.6808837}}.

\bibitem{Ryan1994}
Harold Ryan.
\newblock Ricker, ormsby, klander, butterworth – a choice of wavelets | cseg recorder, 9 1994.
\newblock URL: \url{https://csegrecorder.com/articles/view/ricker-ormsby-klander-butterworth-a-choice-of-wavelets}.

\bibitem{Simonyan15}
Karen Simonyan and Andrew Zisserman.
\newblock Very deep convolutional networks for large-scale image recognition.
\newblock In {\em International Conference on Learning Representations}, 2015.

\bibitem{ILSVRC15}
Olga Russakovsky, Jia Deng, Hao Su, Jonathan Krause, Sanjeev Satheesh, Sean Ma, Zhiheng Huang, Andrej Karpathy, Aditya Khosla, Michael Bernstein, Alexander~C. Berg, and Li~Fei-Fei.
\newblock {ImageNet Large Scale Visual Recognition Challenge}.
\newblock {\em International Journal of Computer Vision (IJCV)}, 115(3):211--252, 2015.
\newblock \href {https://doi.org/10.1007/s11263-015-0816-y} {\path{doi:10.1007/s11263-015-0816-y}}.

\bibitem{Erds1961}
Paul Erdős and Alfréd Rényi.
\newblock On a classical problem of probability theory b.
\newblock 9 1961.
\newblock URL: \url{https://www.renyi.hu/~p_erdos/1961-09.pdf}.

\bibitem{elsken2019neuralarchitecturesearchsurvey}
Thomas Elsken, Jan~Hendrik Metzen, and Frank Hutter.
\newblock Neural architecture search: A survey, 2019.
\newblock URL: \url{https://arxiv.org/abs/1808.05377}, \href {https://arxiv.org/abs/1808.05377} {\path{arXiv:1808.05377}}.

\bibitem{schmidt2019recurrentneuralnetworksrnns}
Robin~M. Schmidt.
\newblock Recurrent neural networks (rnns): A gentle introduction and overview, 2019.
\newblock URL: \url{https://arxiv.org/abs/1912.05911}, \href {https://arxiv.org/abs/1912.05911} {\path{arXiv:1912.05911}}.

\bibitem{Khan_2022}
Salman Khan, Muzammal Naseer, Munawar Hayat, Syed~Waqas Zamir, Fahad~Shahbaz Khan, and Mubarak Shah.
\newblock Transformers in vision: A survey.
\newblock {\em ACM Computing Surveys}, 54(10s):1–41, January 2022.
\newblock URL: \url{http://dx.doi.org/10.1145/3505244}, \href {https://doi.org/10.1145/3505244} {\path{doi:10.1145/3505244}}.

\bibitem{10.5555/645530.655813}
John~D. Lafferty, Andrew McCallum, and Fernando C.~N. Pereira.
\newblock Conditional random fields: Probabilistic models for segmenting and labeling sequence data.
\newblock In {\em Proceedings of the Eighteenth International Conference on Machine Learning}, ICML '01, page 282–289, San Francisco, CA, USA, 2001. Morgan Kaufmann Publishers Inc.

\bibitem{10780702}
Yuning Wang, Zhongqi Yang, Iman Azimi, Amir~M. Rahmani, and Pasi Liljeberg.
\newblock Attention-based explainable ai for wearable multivariate data: A case study on affect status prediction.
\newblock In {\em 2024 IEEE 20th International Conference on Body Sensor Networks (BSN)}, pages 1--4, 2024.
\newblock \href {https://doi.org/10.1109/BSN63547.2024.10780702} {\path{doi:10.1109/BSN63547.2024.10780702}}.

\end{thebibliography}
\end{document}